\documentclass[11pt]{article}

\usepackage{acl}

\usepackage{times}
\usepackage{latexsym}

\usepackage[T1]{fontenc}

\usepackage[utf8]{inputenc}

\usepackage{microtype}

\usepackage{inconsolata}

\usepackage{graphicx}
\usepackage{amsmath}
\usepackage{algorithm}
\usepackage{algpseudocode}
\usepackage{amsfonts}
\usepackage{booktabs}
\usepackage{multirow}
\usepackage{multicol}
\usepackage[table]{xcolor}

\definecolor{groupgray}{HTML}{F2F2F2}  
\definecolor{topblue}{HTML}{D6EAF8}   
\definecolor{toporange}{HTML}{FDEBD0}
\usepackage{threeparttable}
\usepackage{float}   
\usepackage{subcaption} 
\usepackage{caption}
\usepackage{booktabs} 
\usepackage{tabularx} 
\usepackage{booktabs} 
\usepackage{cuted}
\usepackage{enumitem}

\title{YEZE at SemEval-2026 Task 9: Detecting Multilingual, Multicultural and Multievent Online Polarization via Heterogeneous Ensembling}

\author{Fengze Guo \quad Yue Chang\\
    University of Tübingen \\
    \{fengze.guo, yue.chang\}@student.uni-tuebingen.de}

\begin{document}
\maketitle

\begin{abstract}
This paper presents our system for \textit{SemEval-2026 Task 9: Detecting Multilingual, Multicultural and Multievent Online Polarization}, which identifies polarized social media content in 22 languages through three subtasks: binary detection, target classification, and manifestation identification. We propose a heterogeneous ensemble of multilingual pretrained models, combining XLM-RoBERTa-large and mDeBERTa-v3-base. We investigate techniques such as multi-task learning, translation-based data augmentation, and class weighting to improve classification performance under severe label imbalance. Our findings indicate that independent task modeling combined with class weighting is more effective.
\end{abstract}


\section{Introduction}
Social media platforms have expanded public communication worldwide, but they have also amplified online polarization across languages and cultures \cite{howard2019free}. Polarized content differs not only in targets and stance but also in its linguistic manifestations, making automatic identification important for large-scale analysis of social dynamics and for assisting downstream moderation and policy research.

This paper presents our system for \textit{SemEval-2026 Task 9}~\cite{naseem-etal-2026-polar}, a shared task dedicated to identifying online polarization in multilingual social media.\footnote{The code is available at \url{https://github.com/FezeGo/SemEval-2026-Task9-Polar}.} The shared task comprises three subtasks: (1) binary classification to determine whether a post is polarized, (2) multi-label classification of the polarization target type, and (3) multi-label identification of manifestation categories. The official dataset~\cite{naseem2026polarbenchmarkmultilingualmulticultural} consists of multilingual social media posts in 22 languages: Amharic, Arabic, Bengali, Burmese, Chinese, English, German, Hausa, Hindi, Italian, Khmer, Nepali, Odia, Persian, Polish, Punjabi, Russian, Spanish, Swahili, Telugu, Turkish, and Urdu. Subtasks~1 and~2 cover all 22 languages, while Subtask~3 excludes Italian, Polish, Russian, and Burmese. Detailed input/output schemas and language coverage are provided in Appendices~\ref{app:inout} and~\ref{app:lang}. All subtasks are evaluated using the macro-averaged F1 score (Macro-F1).

Our system covers all subtasks, modeling each as an independent problem. Our final system combines two complementary multilingual encoders (XLM-RoBERTa-large~\cite{conneau-etal-2020-unsupervised} and mDeBERTa-v3-base~\cite{mdeberta}) in a weighted ensemble optimized on the development set. For the multi-label subtasks, we use binary relevance with weighted binary cross-entropy to mitigate severe label imbalance.

This task poses several challenges for multilingual modeling. We observe substantial cross-lingual variation in label distributions, with strong prior shift in Subtask~1 and pronounced sparsity in Subtasks~2 and~3, making minority-label learning and calibration central under Macro-F1. Our experiments also suggest that shared multi-task training can introduce negative transfer when sparse fine-grained labels compete with the dominant binary objective, making fine-grained prediction less stable than coarse binary detection. Together, these observations indicate that fine-grained label sparsity, cross-task inconsistency, and negative transfer remain the main bottlenecks, especially for manifestation prediction.

In the official evaluation, we ranked in the Top~10 for 11/22, 16/22, and 17/18 languages in Subtasks~1,~2, and~3, respectively. The detailed per-language rankings are presented in Appendix \ref{app:rankings}.


\section{Background}
Online polarization is an intricate socio-technical phenomenon. Beyond algorithmic amplification within ``information bubbles''~\cite{buildup2025polarization}, social network structures often foster echo chambers where bipolar discourse is costly and infrequent~\cite{garimella2018politicaldiscoursesocialmedia}. Crucially, research suggests that mere exposure to opposing viewpoints can paradoxically increase polarization~\cite{Bailsocialmedia}, underscoring the deep-seated nature of digital rifts. These divisions have severe offline consequences, such as intensifying ethnic mobilization and marginalizing vulnerable voices in conflict zones~\cite{AliYimamAyeleBiemannSemmann}. In NLP, early work largely focused on identifying predictive features for hate speech and offensive content~\cite{waseem-hovy-2016-hateful,davidson2017automated}. Recent frameworks like HateCheck~\cite{rottger-etal-2021-hatecheck} have introduced functional testing to evaluate model robustness against complex linguistic phenomena such as negation and counter-speech. 

Recent benchmarks~\cite{naseem2026polarbenchmarkmultilingualmulticultural} suggest that while binary detection is relatively mature, fine-grained target and manifestation prediction remains substantially more challenging. In addition, LLM-based approaches have been shown to be less reliable than specialized supervised models for these fine-grained labels. Motivated by these limitations, we focus on supervised multilingual encoder-based models, which provide more stable and controllable optimization in low-resource and imbalanced settings.

Such models are typically based on Transformer architectures~\cite{vaswani2017attention}, with pre-trained variants such as BERT~\cite{devlin2018bert} offering strong contextual representations for downstream classification tasks. 

Within this paradigm, a common strategy is to jointly model related objectives using multi-task learning (MTL)~\cite{caruana1997multitask}. However, prior work has shown that MTL can suffer from negative transfer and task interference, particularly when label distributions are highly imbalanced or heterogeneous~\cite{yu2020gradient}. In contrast, we adopt independent task modeling combined with ensemble learning, which proves to be more robust under severe label sparsity.


\section{System Overview}

Our system targets multilingual polarization modeling across 22 languages under severe label imbalance. We follow three core design principles: (i) per-subtask specialization to ensure high accuracy on each individual task, (ii) imbalance-aware optimization to tackle class imbalance in multi-label subtasks, and (iii) heterogeneous ensembling to enhance robustness across languages and events.

\subsection{Model Architecture}
We treat the three subtasks as independent classification problems, with separate models trained for each. Our architecture leverages two complementary multilingual encoders to maximize performance across diverse languages and subtasks:

\paragraph{XLM-RoBERTa-large.}
We select XLM-RoBERTa-large (XLM-R) as the primary backbone because it is a strong multilingual encoder with broad cross-lingual coverage and stable transfer performance in prior work~\cite{conneau-etal-2020-unsupervised}.

\paragraph{mDeBERTa-v3-base.}
We also employ mDeBERTa-v3-base (mDeBERTa), which uses disentangled attention to encode content and relative position separately~\cite{mdeberta}.

Together, these two encoders provide complementary representation spaces and tokenization behaviors, which are then combined through ensembling to increase robustness across different languages and tasks.

\subsection{Imbalance-aware Optimization}
\begin{figure*}[t]
    \centering
    \includegraphics[width=0.8\textwidth]{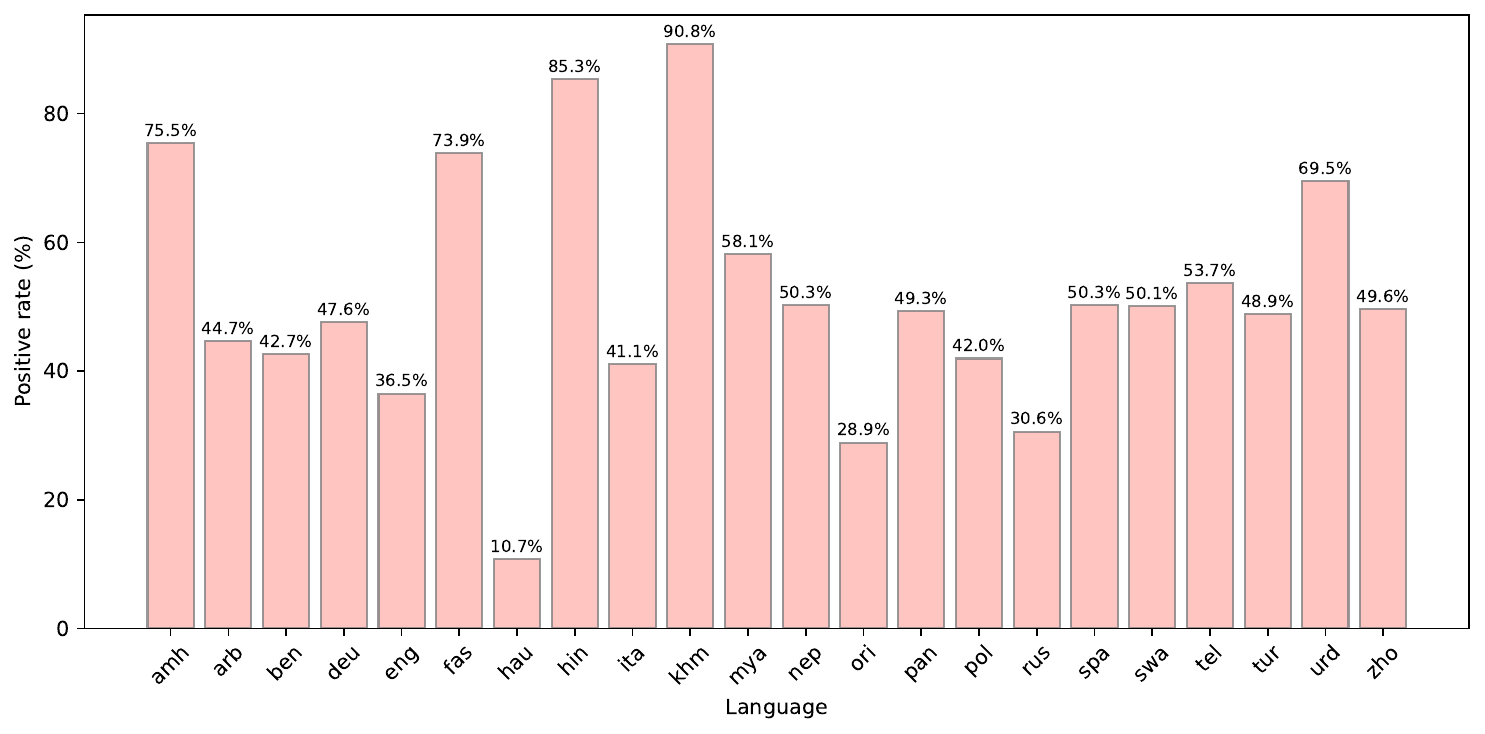}
    \caption{Positive rates for Subtask~1 across languages in the merged train+dev set.}
    \label{fig:s1-prior-shift}
\end{figure*}

As shown in Figure~\ref{fig:s1-prior-shift}, the polarized rate varies substantially across languages, indicating strong cross-lingual prior shift. This motivates the use of imbalance-aware optimization, especially for the multi-label subtasks where sparsity is even more severe.

For Subtasks~2 and~3, we adopt a Binary Relevance (BR) formulation~\cite{read2011classifier}, where multi-label predictions are decomposed into independent binary classification tasks. This allows each label to be treated as a separate problem, with a sigmoid activation applied to each logit to yield independent probabilities:
\[
\hat{y}_c = P(c=1 \mid x) = \sigma(z_c) \in [0,1].
\]

To handle severe class imbalance, we utilize Weighted Binary Cross-Entropy (WBCE) loss, a common approach in cost-sensitive and imbalanced learning~\cite{Elkan2001TheFO, he2009learning}, where the weight for each class is inversely proportional to the frequency of its occurrences in the training data:

\begin{equation*}
\begin{split}
\mathcal{L}(y_c, \hat{y}_c) = - [ & w_c \cdot y_c \log(\hat{y}_c) + \\
& (1 - y_c) \log(1 - \hat{y}_c) ],
\end{split}
\end{equation*}
where the weight $w_c$ is computed as:
\[
w_c = \frac{N_{\text{neg},c}}{\max(N_{\text{pos},c}, 1)}.
\]
Here, $N_{\text{neg},c}$ and $N_{\text{pos},c}$ represent the number of negative and positive instances for each label, respectively.

We also evaluated the effectiveness of WBCE relative to other loss functions; a detailed comparison is provided in Section~\ref{sec:experiments}.

\subsection{Heterogeneous Ensemble Strategy}
For our final submission, we ensemble the two backbones using weighted probability averaging. For each label $c$ and input post $x$, the ensembled posterior is:
\begin{equation*}
\begin{split}
\bar{P}(c =1 \mid x) = & \alpha \, P_{\text{XLM-R}}(c =1 \mid x)+
\\
&  (1-\alpha)\, P_{\text{mDeBERTa}}(c =1 \mid x),
\end{split}
\end{equation*}
where $\alpha$ is selected on the \textit{dev} set to maximize Macro-F1 (we use $\alpha=0.7$).

\subsection{Pipeline Summary}
Our training and inference pipeline is:
\begin{enumerate}[label = (\roman*), nosep]
    \item fine-tune XLM-R and mDeBERTa separately for each subtask,
    \item tune the ensemble weight $\alpha$ on the \textit{dev} set,
    \item generate test predictions by weighted ensembling and a global threshold $\tau=0.5$.
\end{enumerate}
Experimental details are provided in Section~\ref{sec:experimentsetup}.

\subsection{System Variants}
We evaluate: 
\begin{enumerate}[label = (\alph*), nosep]
    \item single-backbone models (XLM-R or mDeBERTa),
    \item an MTL variant with a shared XLM-R encoder and task-specific heads,
    \item the heterogeneous ensemble with independent per-subtask training (submitted).
\end{enumerate}

\begin{table*}[t]
\centering
\small
\setlength{\tabcolsep}{3pt}
\renewcommand{\arraystretch}{0.92}
\begin{tabular}{l | cccc | cccc | cccc}
\toprule
\multirow{2}{*}{\textbf{Lang.}} & \multicolumn{4}{c|}{\textbf{POLARDETECT (Subtask~1)}} & \multicolumn{4}{c|}{\textbf{POLARTYPE (Subtask~2)}} & \multicolumn{4}{c}{\textbf{POLARMANIFEST (Subtask~3)}} \\
 & XLM-R & mDeB & \textbf{Ensem} & MTL
 & XLM-R & mDeB & \textbf{Ensem} & MTL
 & XLM-R & mDeB & \textbf{Ensem} & MTL \\
\midrule

\rowcolor{groupgray} \multicolumn{13}{l}{\textbf{Indo-Aryan Family}} \\
hin & 0.801 & 0.799 & \cellcolor{topblue}\textbf{0.806} & 0.785 & 0.764 & 0.747 & \cellcolor{topblue}\textbf{0.776} & 0.726 & 0.737 & 0.728 & \cellcolor{topblue}\textbf{0.741} & 0.732 \\
urd & 0.785 & 0.748 & 0.789 & \cellcolor{toporange}\textbf{0.794} & 0.779 & 0.770 & 0.781 & \cellcolor{toporange}\textbf{0.785} & 0.811 & 0.804 & \cellcolor{topblue}\textbf{0.815} & 0.815 \\
ben & 0.840 & 0.836 & \cellcolor{topblue}\textbf{0.840} & 0.838 & 0.351 & \cellcolor{toporange}\textbf{0.375} & 0.352 & 0.350 & 0.195 & \cellcolor{toporange}\textbf{0.208} & 0.202 & 0.186 \\
nep & 0.905 & 0.888 & 0.908 & \cellcolor{toporange}\textbf{0.909} & 0.731 & 0.747 & \cellcolor{topblue}\textbf{0.750} & 0.721 & 0.649 & 0.577 & \cellcolor{topblue}\textbf{0.653} & 0.629 \\
pan & 0.777 & 0.773 & \cellcolor{topblue}\textbf{0.786} & 0.780 & 0.414 & 0.401 & \cellcolor{topblue}\textbf{0.421} & 0.382 & 0.494 & 0.498 & \cellcolor{topblue}\textbf{0.504} & 0.491 \\
ori & \cellcolor{toporange}\textbf{0.818} & 0.743 & 0.812 & 0.792 & \cellcolor{toporange}\textbf{0.568} & 0.443 & 0.562 & 0.556 & \cellcolor{toporange}\textbf{0.301} & 0.235 & 0.277 & 0.243 \\

\midrule
\rowcolor{groupgray} \multicolumn{13}{l}{\textbf{European}} \\
eng & 0.761 & 0.787 & 0.781 & \cellcolor{toporange}\textbf{0.792} & 0.489 & 0.484 & \cellcolor{topblue}\textbf{0.503} & 0.481 & \cellcolor{toporange}\textbf{0.501} & 0.483 & 0.500 & 0.499 \\
deu & \cellcolor{toporange}\textbf{0.722} & 0.685 & 0.717 & 0.711 & 0.567 & 0.559 & \cellcolor{topblue}\textbf{0.573} & 0.561 & 0.487 & 0.485 & 0.493 & \cellcolor{toporange}\textbf{0.494} \\
spa & 0.774 & 0.770 & 0.783 & \cellcolor{toporange}\textbf{0.784} & 0.626 & 0.621 & \cellcolor{topblue}\textbf{0.637} & 0.631 & \cellcolor{toporange}\textbf{0.506} & 0.468 & 0.505 & 0.499 \\
ita & 0.525 & \cellcolor{toporange}\textbf{0.662} & 0.591 & 0.548 & \cellcolor{toporange}\textbf{0.323} & 0.293 & 0.319 & 0.310 & -- & -- & -- & -- \\
rus & 0.764 & 0.772 & \cellcolor{topblue}\textbf{0.806} & 0.794 & \cellcolor{toporange}\textbf{0.568} & 0.516 & 0.558 & 0.567 & -- & -- & -- & -- \\
pol & 0.798 & 0.786 & 0.810 & \cellcolor{toporange}\textbf{0.819} & 0.557 & 0.508 & \cellcolor{topblue}\textbf{0.578} & 0.567 & -- & -- & -- & -- \\

\midrule
\rowcolor{groupgray} \multicolumn{13}{l}{\textbf{Afro-Asiatic \& Iranic}} \\
arb & 0.814 & 0.818 & 0.819 & \cellcolor{toporange}\textbf{0.830} & 0.614 & 0.604 & \cellcolor{topblue}\textbf{0.626} & 0.600 & 0.608 & 0.591 & \cellcolor{topblue}\textbf{0.610} & 0.596 \\
fas & 0.790 & 0.794 & \cellcolor{topblue}\textbf{0.814} & 0.780 & 0.569 & 0.576 & \cellcolor{topblue}\textbf{0.580} & 0.552 & \cellcolor{toporange}\textbf{0.414} & 0.365 & 0.400 & 0.400 \\
amh & 0.782 & 0.757 & 0.771 & \cellcolor{toporange}\textbf{0.784} & 0.608 & 0.629 & \cellcolor{topblue}\textbf{0.649} & 0.631 & 0.544 & 0.510 & \cellcolor{topblue}\textbf{0.546} & 0.522 \\
hau & \cellcolor{toporange}\textbf{0.774} & 0.686 & 0.740 & \cellcolor{toporange}\textbf{0.774} & 0.375 & 0.318 & \cellcolor{topblue}\textbf{0.381} & 0.343 & 0.122 & 0.115 & 0.122 & \cellcolor{toporange}\textbf{0.187} \\

\midrule
\rowcolor{groupgray} \multicolumn{13}{l}{\textbf{Other Families}} \\
zho & 0.896 & 0.867 & \cellcolor{topblue}\textbf{0.897} & 0.892 & 0.771 & 0.723 & \cellcolor{topblue}\textbf{0.782} & 0.741 & 0.555 & 0.455 & \cellcolor{topblue}\textbf{0.568} & 0.540 \\
mya & 0.862 & 0.874 & \cellcolor{topblue}\textbf{0.879} & 0.871 & 0.672 & 0.672 & \cellcolor{topblue}\textbf{0.692} & 0.619 & -- & -- & -- & -- \\
khm & \cellcolor{toporange}\textbf{0.742} & 0.673 & 0.727 & 0.718 & 0.557 & \cellcolor{toporange}\textbf{0.675} & 0.667 & 0.644 & \cellcolor{toporange}\textbf{0.361} & 0.283 & 0.353 & 0.341 \\
tel & 0.880 & 0.860 & \cellcolor{topblue}\textbf{0.884} & 0.860 & 0.419 & 0.414 & \cellcolor{topblue}\textbf{0.429} & 0.423 & \cellcolor{toporange}\textbf{0.389} & 0.369 & 0.378 & 0.350 \\
tur & 0.761 & 0.765 & 0.777 & \cellcolor{toporange}\textbf{0.791} & 0.556 & 0.549 & \cellcolor{topblue}\textbf{0.563} & 0.558 & 0.504 & 0.489 & 0.500 & \cellcolor{toporange}\textbf{0.506} \\
swa & 0.763 & 0.778 & 0.778 & \cellcolor{toporange}\textbf{0.784} & 0.460 & \cellcolor{toporange}\textbf{0.484} & 0.483 & 0.433 & 0.547 & 0.536 & \cellcolor{topblue}\textbf{0.547} & 0.540 \\

\midrule
\rowcolor{groupgray} \textbf{Average} & 0.788 & 0.778 & \cellcolor{topblue}\textbf{0.796} & 0.792
& 0.565 & 0.550 & \cellcolor{topblue}\textbf{0.575} & 0.554
& \cellcolor{toporange}\textbf{0.485} & 0.456 & 0.484 & 0.476 \\
\bottomrule
\end{tabular}
\caption{Official Macro-F1 results by language for Subtasks 1--3. \textbf{Ensem} is our final submission; \textit{mDeB} denotes mDeBERTa-v3-base. \colorbox[HTML]{E0F0FF}{blue} indicates that the ensemble is tied for or achieves the best score, while \colorbox[HTML]{FFF0E0}{orange} indicates that the best non-ensemble model achieves a strictly higher score than the ensemble.}
\label{tab:final-table}
\end{table*}


\section{Experimental Setup}
\label{sec:experimentsetup}
\subsection{Data and Partitioning}
\label{sec:data}

We restrict training to the provided task data and do not use external lexicons. The official dataset~\cite{naseem2026polarbenchmarkmultilingualmulticultural} provides \textit{train}, \textit{dev}, and \textit{test} partitions.

\paragraph{Development Phase.} We used an internal 85/15 split of the official \textit{train} set for model selection. For Subtask~1, standard stratification was applied; for Subtasks~2 and~3, we used iterative stratification~\cite{Sechidis2011OnTS} to preserve rare label co-occurrences under extreme sparsity.
\paragraph{Test Phase.} Upon the release of \textit{dev} labels, we utilized the official \textit{dev} set as a hold-out validation set for final hyperparameter tuning. The final system was retrained on the union of the official \textit{train} and \textit{dev} sets to generate predictions for the hidden \textit{test} set.

\subsection{Evaluation Metrics}
All subtasks are evaluated using Macro-F1:
\[
\text{Macro-F1} = \frac{1}{|L|} \sum_{l \in L} F1_l.
\]
For Subtasks~2 and~3, $F1_l$ is computed independently for each label and then averaged (label-wise macro), which emphasizes rare categories.

\subsection{Implementation}

\paragraph{Preprocessing.}
We applied minimal preprocessing: rows with missing or empty text were removed, while emojis were retained to preserve affective cues. No task-specific normalization beyond shared-task standardization (e.g., URLs as \texttt{[URL]}) was performed. Tokenization used the default tokenizer with truncation and dynamic padding, and a maximum sequence length of 256.

\paragraph{Training setup.}
Experiments were conducted on NVIDIA A100 GPUs using PyTorch and Hugging Face Transformers (\texttt{bf16}/\texttt{tf32}). Models were fine-tuned with AdamW and a linear scheduler (warmup ratio 0.1) with weight decay set to 0. On the internal validation set, we use early stopping with patience 2. The learning rate was $1\times10^{-5}$ for XLM-R (epochs 4, batch size 32) and $2\times10^{-5}$ for mDeBERTa (epochs 5, batch size 64). Additional implementation details are given in Appendix~\ref{app:imple}.

\paragraph{Thresholds and ensembling.}
For multi-label settings, weighted binary cross-entropy was used with a positive weight computed from label frequencies. Per-label threshold tuning was not adopted due to performance degradation, so a global threshold $\tau=0.5$ was used. The ensemble coefficient $\alpha$ was set to 0.7 after grid search.


\section{Results and Discussion}
\label{sec:analysis}

\begin{figure*}[t]
  \centering
  \includegraphics[width=\textwidth]{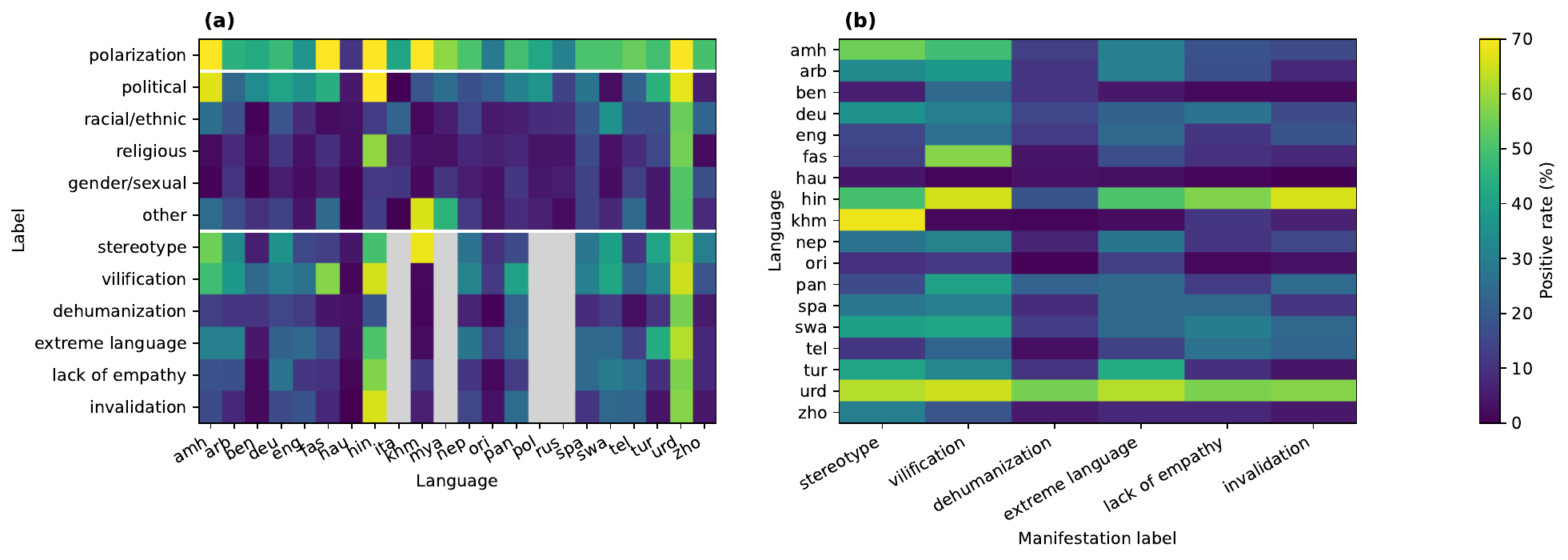}
  \caption{Dataset imbalance analysis on the merged train+dev set. (a) Task-level skew (Language $\times$ Task). (b) Subtask~3 label imbalance (Language $\times$ Label).}
  \label{fig:imbalance}
\end{figure*}

\subsection{Official Evaluation Results}
\label{sec:ranking}

Table~\ref{tab:final-table} reports official test Macro-F1 by language for all subtasks. We use the ensemble as the final submission across all subtasks to maintain a single consistent pipeline. In Subtask~2, it reaches Top~5 in Amharic~(4th), Urdu~(5th), Odia~(5th), and Polish~(5th). In Subtask~3, it ranks in the Top~5 for Amharic~(4th), Arabic~(3rd), English~(5th), Khmer~(5th), Spanish~(4th), and Urdu~(3rd).

\subsection{Data Imbalance and Cross-lingual Conditions}
\label{sec:imbalance}
Figure~\ref{fig:imbalance} summarizes train+dev label distributions across subtasks and languages. Figure~\ref{fig:imbalance}(a) shows strong cross-lingual prior shift in Subtask~1 (10.75/49.45/90.79\% min/med/max polarized rate) and overall skew across subtasks. As shown in Figure~\ref{fig:imbalance}(b), \textit{Vilification} and \textit{Stereotype} are more frequent than the remaining manifestation labels. Overall, difficulty stems from sparsity interacting with cross-lingual heterogeneity~\cite{hu2020xtrememassivelymultilingualmultitask}. Despite partial intra-family consistency~\cite{pires-etal-2019-multilingual,wu-dredze-2019-beto}, extreme sparsity~\cite{lauscher-etal-2020-zero} and script/tokenization effects~\cite{rust-etal-2021-good} still limit fine-grained transfer.

\subsection{Optimization Analysis}
\label{sec:experiments}

\paragraph{Loss functions.}
Development tests (Table~\ref{tab:ablation-loss}) show Focal Loss \cite{focal-loss} benefits well-represented languages but is unstable under sparsity. Conversely, WBCE significantly boosts low-resource settings (e.g., yielding a +0.212 increase in Macro-F1 for Telugu) and improves the 22-language average by 3.5 points (0.492$\rightarrow$0.527).

\begin{table}[htbp]
\centering
\renewcommand{\arraystretch}{1.3}
\setlength{\tabcolsep}{10pt}   
\small                   
\begin{tabular}{lcccc}
\toprule
\textbf{Lang.} & \textbf{Base} & \textbf{Focal} & \textbf{WBCE} & \textbf{$\Delta$} \\ \midrule
zho & 0.6905 & 0.6893 & \textbf{0.7218} & +0.0313 \\
hin & 0.7724 & \textbf{0.8127} & 0.7996 & +0.0272 \\
tel & 0.2253 & 0.2986 & \textbf{0.4372} & +0.2119 \\
amh & 0.3760 & 0.4568 & \textbf{0.4589} & +0.0829 \\
hau & 0.1115 & \textbf{0.2719} & 0.2513 & +0.1398 \\ \midrule
\rowcolor[HTML]{F2F2F2}
\textbf{Avg.} & 0.4351 & 0.5053 & \textbf{0.5338} & +0.0987 \\ \bottomrule
\end{tabular}
\caption{Optimization objective ablation for Subtask~2 (Macro-F1). \textbf{Base} = BCE loss; \textbf{Focal} = Focal Loss; \textbf{$\Delta$} = the improvement of WBCE over Base.}
\label{tab:ablation-loss}
\end{table}

\paragraph{MTL vs. Independent Modeling.}
The multi-task learning (MTL) variant suggests negative transfer for the sparsest labels. While shared representations can slightly benefit Subtask~1, Subtasks~2 and~3 suffer when low-frequency labels compete with the dominant binary objective during shared updates~\cite{yu2020gradient}. We use independent per-subtask modeling to preserve fine-grained signals.

\subsection{Error Analysis and Post-hoc Evaluation}
\label{sec:error}
Post-hoc evaluation confirms strong cross-lingual variability; additional figures are shown in Appendix~\ref{app:posthoc}. Weak transfer from Subtask~1 to Subtasks~2 and~3 (e.g., Bengali and Telugu) highlights fine-grained label sparsity. Category-wise, \textit{Political} (Subtask~2), \textit{Vilification} and \textit{Stereotype} (Subtask~3) are the most reliable, while \textit{Gender/Sexual}, \textit{Religious}, and \textit{Other} (Subtask~2), as well as \textit{Dehumanization}, \textit{Lack of Empathy}, and \textit{Invalidation} (Subtask~3), remain challenging.

\paragraph{Miscalibration and label collapse.}
Low-resource settings exhibit miscalibration (high recall but low precision) and label collapse, where sparse categories fall to near-zero F1, suggesting expanded but noisy decision regions~\cite{desai-durrett-2020-calibration}.

\paragraph{Cross-task inconsistency.}
Independent modeling causes hierarchical violations: some instances labeled non-polarized in Subtask~1 receive positive Subtask~2 or~3 labels. This suggests a trade-off between robustness and coherence, motivating future post-hoc gating or hierarchical calibration.

\paragraph{Translation augmentation.}
Machine translation augmentation (using 4,000 Gemini-generated samples for Hausa) failed to yield significant performance gains. This result suggests that synthetic ``translationese'' and the resulting pragmatic shifts~\cite{ji2023cultural} may obscure region-specific idioms and subtle linguistic cues.


\section{Conclusion}

We presented a multilingual system for SemEval-2026 Task 9 that addresses polarization detection and characterization across 22 languages.
Our final system combines XLM-R and mDeBERTa in a heterogeneous ensemble and uses imbalance-aware optimization to improve robustness.
Across official evaluations, the system remains competitive and shows that independent per-subtask modeling is a strong practical alternative to shared MTL.

Our analyses show that the main challenges are not only cross-lingual variation, but also fine-grained label sparsity, calibration instability, and cross-task inconsistency.
Post-hoc evaluation on the released test gold labels further confirms that high-coverage labels are substantially easier to learn, while sparse manifestation labels remain the most fragile.
In future work, we plan to explore lightweight hierarchical calibration (e.g., gating-based consistency constraints) and culturally grounded data synthesis to better capture region-specific socio-political idioms for fine-grained polarization analysis.


\section*{Acknowledgments}

We would like to express our sincere gratitude to Dr. Çağrı Çöltekin for his invaluable guidance, insightful comments, and continuous support throughout this study.

\bibliography{custom}

\newpage

\appendix
\onecolumn
\section{Input/Output Schemas}
\label{app:inout}
A subset of the training data for each subtask has been selected as illustrative examples. The corresponding input and output schemas for the three subtasks are presented below.
\subsection{Subtask~1}
This subtask comprises a binary classification task to determine whether a given text exhibits polarized content. A text is labeled as polarized (\textit{Polarization} = 1) only if it explicitly expresses an opinion indicative of attitude polarization, accounting for the overall context and semantic meaning. Texts lacking such characteristics are labeled as non-polarized (\textit{Polarization} = 0).

\begin{table}[H] 
    \centering
    \small 
    \begin{tabularx}{\textwidth}{l X c}
        \toprule
        \textbf{ID} & \textbf{Text} & \textbf{Polarization} \\
        \midrule
        eng\_9739... & Is detecting imperialism in the dnd chat. & -- \\
        eng\_891d... & Start by not listening to msnbc. & -- \\
        \bottomrule
    \end{tabularx}
    \caption{Subtask~1 input examples.}
    \label{tab:s1-input}
\end{table}

\begin{table}[H]
    \centering
    \small
    \begin{tabular}{lc}
        \toprule
        \textbf{ID} & \textbf{Polarization} \\
        \midrule
        eng\_9739... & 0 \\
        eng\_891d... & 1 \\
        \bottomrule
    \end{tabular}
    \caption{Subtask~1 output examples.}
    \label{tab:s1-output}
\end{table}


\subsection{Subtask~2}
This multi-label classification task identifies the specific targets of polarization in each text. Each text is evaluated across five predefined categories: \textit{political}, \textit{racial/ethnic}, \textit{religious}, \textit{gender/sexual}, and \textit{other}. A label of 1 denotes the presence of a given polarization type, while 0 indicates its absence. Multiple labels may be assigned to a single text if multiple types are present.
\begin{table}[H]
    \centering
    \small
    \begin{tabularx}{\columnwidth}{l X c c c c c}
        \toprule
        \textbf{ID} & \textbf{Text} & \textbf{Political} & \textbf{Racial/Ethnic} & \textbf{Religious} & \textbf{Gender/Sexual} & \textbf{Other}\\
        \midrule
        eng\_9739... & Is detecting imperialism in the dnd chat. & -- & -- & -- & -- & -- \\
        eng\_891d... & Start by not listening to msnbc. & -- & -- & -- & -- & -- \\
        \bottomrule
    \end{tabularx}
    \caption{Subtask~2 input examples.}
    \label{tab:s2-input}
\end{table}

\begin{table}[H]
    \centering
    \small
    \begin{tabular}{lccccc}
        \toprule
        \textbf{ID} & \textbf{Political} & \textbf{Racial/Ethnic} & \textbf{Religious} & \textbf{Gender/Sexual} & \textbf{Other}\\
        \midrule
        eng\_9739... & 0 & 0 & 0 & 0 & 0\\
        eng\_891d... & 1 & 0 & 0 & 0 & 1\\
        \bottomrule
    \end{tabular}
    \caption{Subtask~2 output examples.}
    \label{tab:s2-output}
\end{table}


\subsection{Subtask~3}
This subtask identifies the specific manifestations of polarization in each text. Each text is evaluated across six predefined categories: \textit{Stereotype}, \textit{Vilification}, \textit{Dehumanization}, \textit{Extreme\_Language}, \textit{Lack\_of\_Empathy}, and \textit{Invalidation}. A label of 1 denotes presence, while 0 indicates absence. Multiple labels may be assigned to a single text if multiple manifestations are present. 

For brevity, texts and category names are abbreviated in the input table.

\begin{table}[H]
    \centering
    \small 
    \begin{tabularx}{\columnwidth}{l X c c c c c c } 
        \toprule
        \textbf{ID} & \textbf{Text} & \textbf{Stereo.} & \textbf{Vilifi.} & \textbf{Dehuman.} & \textbf{Extreme\_Language} & \textbf{Lack\_of\_Empathy} & \textbf{Invalidation} \\
        \midrule
        eng\_9739... & Is... & -- & -- & -- & -- & -- & -- \\
        eng\_891d... & Start... & -- & -- & -- & -- & -- & -- \\
        \bottomrule
    \end{tabularx}
    \caption{Subtask~3 input examples. Abbreviations: Stereo.=Stereotype; Vilifi.=Vilification; 
Dehuman.=Dehumanization.}
    \label{tab:s3-input}
\end{table}

\begin{table}[H]
    \centering
    \small
    \begin{tabular}{lcccccc}
        \toprule
        \textbf{ID} & \textbf{Stereotype} & \textbf{Vilification} & \textbf{Dehumanization} & \textbf{Extreme\_Language} & \textbf{Lack\_of\_Empathy} & \textbf{Invalidation} \\
        \midrule
        eng\_9739... & 0 & 0 & 0 & 0 & 0 & 0 \\
        eng\_891d... & 1 & 0 & 0 & 0 & 1 & 0 \\
        \bottomrule
    \end{tabular}
    \caption{Subtask~3 output examples.}
    \label{tab:s3-output}
\end{table}

\newpage

\section{Data Statistics}
\label{app:lang}

Table~\ref{tab:detailed-data} provides the detailed instance counts for each language across the training, development, and test sets.

\begin{table}[ht]
\centering
\small
\renewcommand{\arraystretch}{1.5}
\setlength{\tabcolsep}{10pt}
\begin{tabular}{llrrrr}
\toprule
\textbf{Language} & \textbf{Code} & \textbf{Train} & \textbf{Dev} & \textbf{Test} & \textbf{Total} \\
\midrule
Amharic & amh & 3332 & 166 & 1501 & 4999 \\
Arabic & arb & 3380 & 169 & 1521 & 5070 \\
Bengali & ben & 3333 & 166 & 1501 & 5000 \\
Burmese & mya & 2889 & 144 & 1301 & 4334 \\
Chinese & zho & 4280 & 214 & 1927 & 6421 \\
English & eng & 3222 & 160 & 1452 & 4834 \\
German & deu & 3180 & 159 & 1432 & 4771 \\
Hausa & hau & 3651 & 182 & 1644 & 5477 \\
Hindi & hin & 2744 & 137 & 1236 & 4117 \\
Italian & ita & 3334 & 166 & 1538 & 5038 \\
Khmer & khm & 6640 & 332 & 2988 & 9960 \\
Nepali & nep & 2005 & 100 & 903 & 3008 \\
Odia & ori & 2368 & 118 & 1066 & 3552 \\
Persian & fas & 3295 & 164 & 1484 & 4943 \\
Polish & pol & 2391 & 119 & 1077 & 3587 \\
Punjabi & pan & 1700 & 100 & 809 & 2609 \\
Russian & rus & 3348 & 167 & 1508 & 5023 \\
Spanish & spa & 3305 & 165 & 1488 & 4958 \\
Swahili & swa & 6991 & 349 & 3147 & 10487 \\
Telugu & tel & 2366 & 118 & 1066 & 3550 \\
Turkish & tur & 2364 & 115 & 1093 & 3572 \\
Urdu & urd & 3563 & 177 & 1606 & 5346 \\
\midrule
\textbf{Total} & \textbf{22} & \textbf{67736} & \textbf{3744} & \textbf{33782} & \textbf{105262} \\
\bottomrule
\end{tabular}
\caption{Detailed dataset statistics for all 22 languages.}
\label{tab:detailed-data}
\end{table}
These languages cover a broad range of linguistic and cultural contexts, providing a diverse basis for detecting and analyzing online polarization in multilingual settings.

\clearpage
\onecolumn
\section{Detailed Official Rankings}
\label{app:rankings}

Table~\ref{tab:full_rankings} provides the official system rankings across all three subtasks for the 22 languages included in the dataset. Subtask~3 was evaluated on a subset of 18 languages; instances where a language was not included in that subtask are marked with a dash (---).

\begin{table}[ht]
\small
\centering
\renewcommand{\arraystretch}{1.8}
\setlength{\tabcolsep}{12pt}
\begin{tabular}{lccc}
\hline\textbf{Language} & \textbf{Subtask~1} & \textbf{Subtask~2} & \textbf{Subtask~3} \\ \hline
Amharic (\texttt{amh}) & 13 & 4  & 4 \\
Arabic (\texttt{arb})  & 20 & 8  & 3 \\
Bengali (\texttt{ben}) & 10 & 6  & 11 \\
Burmese (\texttt{mya}) & 7  & 9 & --- \\
Chinese (\texttt{zho}) & 12 & 11 & 10  \\
English (\texttt{eng}) & 34 & 8  & 5 \\
German (\texttt{deu})  & 17 & 6  & 7\\
Hausa (\texttt{hau})   & 24 & 7  & 10  \\
Hindi (\texttt{hin})   & 14 & 10 & 6   \\
Italian (\texttt{ita}) & 20 & 8  & ---  \\
Khmer (\texttt{khm})   & 14 & 9  & 5   \\
Nepali (\texttt{nep})  & 9  & 18 & 6   \\
Odia (\texttt{ori})    & 4  & 5  & 6   \\
Persian (\texttt{fas}) & 8 & 11 & 7   \\
Polish (\texttt{pol})  & 9 & 5  &--- \\
Punjabi (\texttt{pan}) & 8  & 14 & 6   \\
Russian (\texttt{rus}) & 7  & 11 & --- \\
Spanish (\texttt{spa}) & 10 & 10 & 4   \\
Swahili (\texttt{swa}) & 17 & 9 & 6   \\
Telugu (\texttt{tel})  & 10 & 9  & 7   \\
Turkish (\texttt{tur}) & 17 & 17 & 6   \\
Urdu (\texttt{urd})    & 8  & 5  & 3   \\ \hline
\end{tabular}
\caption{Official system rankings per language for Subtask~1, Subtask~2, and Subtask~3.}
\label{tab:full_rankings}
\end{table}

\clearpage
\twocolumn
\section{Implementation Details}
\label{app:imple}

\subsection{Preprocessing}
We apply minimal preprocessing across all subtasks.
Rows with missing text are removed, and examples whose cleaned text becomes empty are discarded.
No task-specific normalization is applied beyond the standardization already present in the shared-task data: URLs are already normalized as \texttt{[URL]}, and the released files we use do not contain user placeholders such as \texttt{@USER}.
We retain emojis because they may carry affective or evaluative cues relevant to polarization.

For tokenization, we use the default tokenizer associated with each backbone.
Texts are encoded with \texttt{truncation=True}, \texttt{padding=False}, and \texttt{max\_length=256}; dynamic padding is applied at batch time.
In practice, most posts are short, so truncation is triggered only for a small fraction of examples.

\subsection{Hyperparameter Search and Training}
All models are implemented in PyTorch using the Hugging Face Transformers library.
We tune the learning rate over $\{1\mathrm{e}{-5}, 2\mathrm{e}{-5}\}$ and use AdamW with a linear scheduler and warmup ratio 0.1.
We set weight decay to 0, early stopping uses patience 2, and gradient clipping is not applied.

XLM-R is trained for 4 epochs, and mDeBERTa for 5 epochs.
We use per-device batch sizes of 32 and 64, respectively, with gradient accumulation only when needed due to memory constraints.
We also test three random seeds (42, 2025, 3072), but do not use seed ensembling because trends are consistent across runs.

For multi-label settings, weighted binary cross-entropy is implemented through \texttt{BCEWithLogitsLoss} with a label-dependent \texttt{pos\_weight} computed from class frequencies in the corresponding training split.
For the final run, \texttt{pos\_weight} is recomputed on the merged train+dev data.
Per-label threshold tuning was explored on development data but was not adopted because it reduced overall Macro-F1 under extreme label sparsity.

\subsection{Final Configuration Summary}
After development-stage tuning, we use the following final configuration for the submitted system:
\begin{itemize}
    \item \textbf{XLM-RoBERTa-large:} batch size 32, 4 epochs, learning rate $1\times10^{-5}$.
    \item \textbf{mDeBERTa-v3-base:} batch size 64, 5 epochs, learning rate $2\times10^{-5}$.
    \item \textbf{Thresholding:} global threshold $\tau=0.5$ for Subtask~1 and all labels in Subtasks~2 and~3.
    \item \textbf{Ensembling:} weighted probability averaging with $\alpha=0.7$.
\end{itemize}

\newpage
\onecolumn
\section{Additional Post-hoc Visualizations}
\label{app:posthoc}
These figures provide supplementary post-hoc analyses on the released test gold labels and complement the discussion in Section~\ref{sec:analysis}. They visualize per-language performance, label-wise F1, and precision--recall gaps that are summarized only briefly in the main text.

\subsection{Per-language Performance Across Subtasks}

\begin{figure}[htbp]
    \centering
    \includegraphics[width=\textwidth]{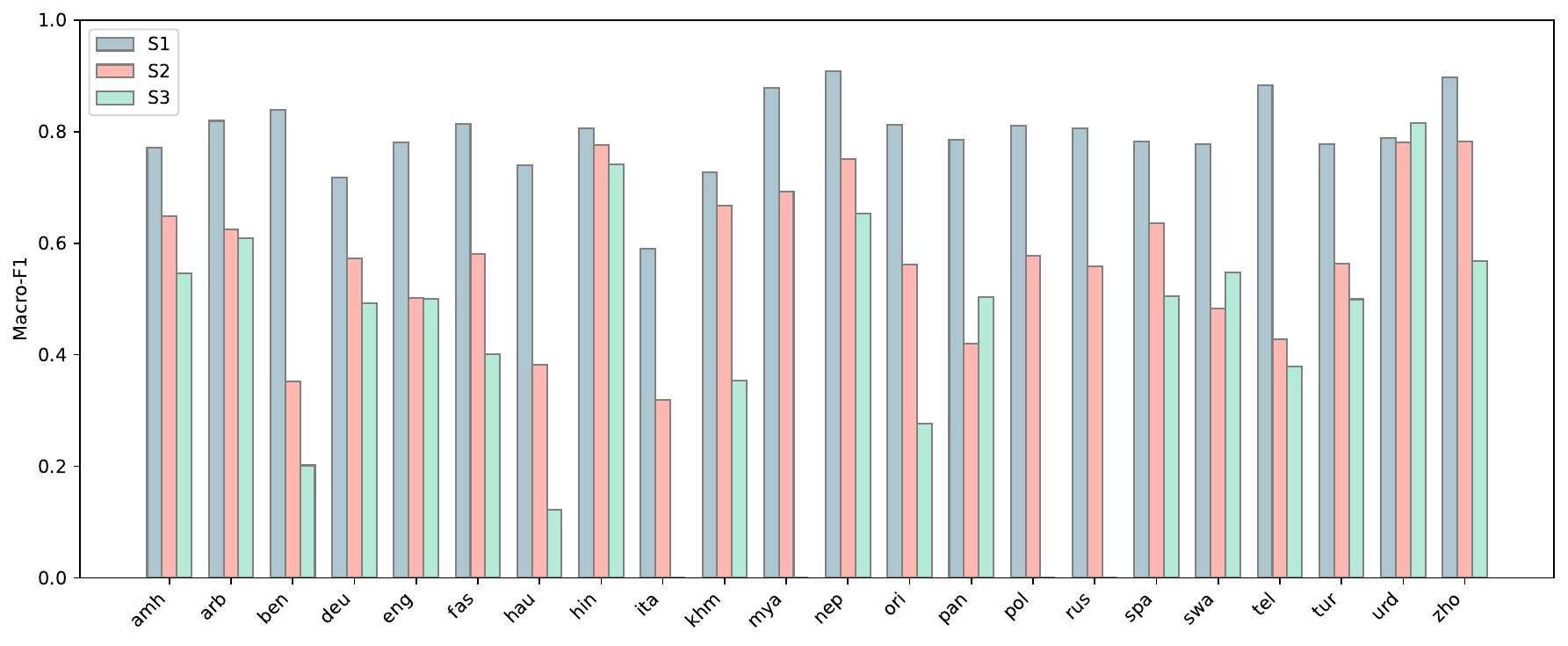}
    \caption{Post-hoc Macro-F1 by language and subtask on the released test gold labels.}
    \label{fig:app-per-lang}
\end{figure}

Figure~\ref{fig:app-per-lang} highlights substantial cross-lingual variability and the mismatch between strong Subtask~1 performance and weaker fine-grained performance in some languages. In particular, several languages remain competitive in binary polarization detection but drop noticeably in Subtasks~2 and~3, which is consistent with the role of label sparsity in the multi-label settings.

\subsection{Additional Analysis for Subtask~2}

\begin{figure}[htbp]
    \centering
    \begin{subfigure}[t]{0.48\textwidth}
        \centering
        \includegraphics[width=\linewidth]{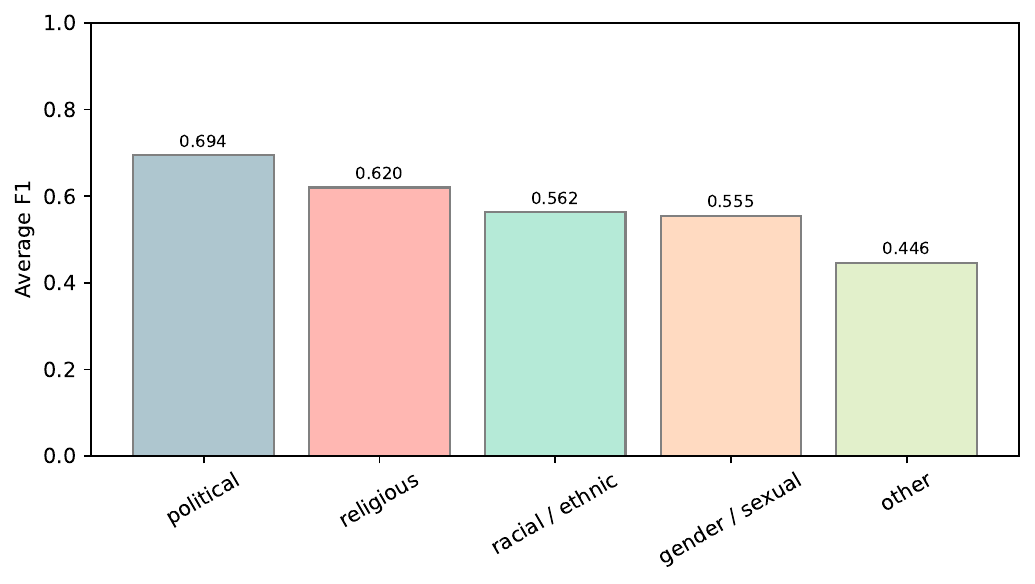}
        \caption{Average label-wise F1 for Subtask~2, showing that some target categories are substantially more stable than others.}
        \label{fig:app-s2-avg-f1}
    \end{subfigure}
    \hfill
    \begin{subfigure}[t]{0.48\textwidth}
        \centering
        \includegraphics[width=\linewidth]{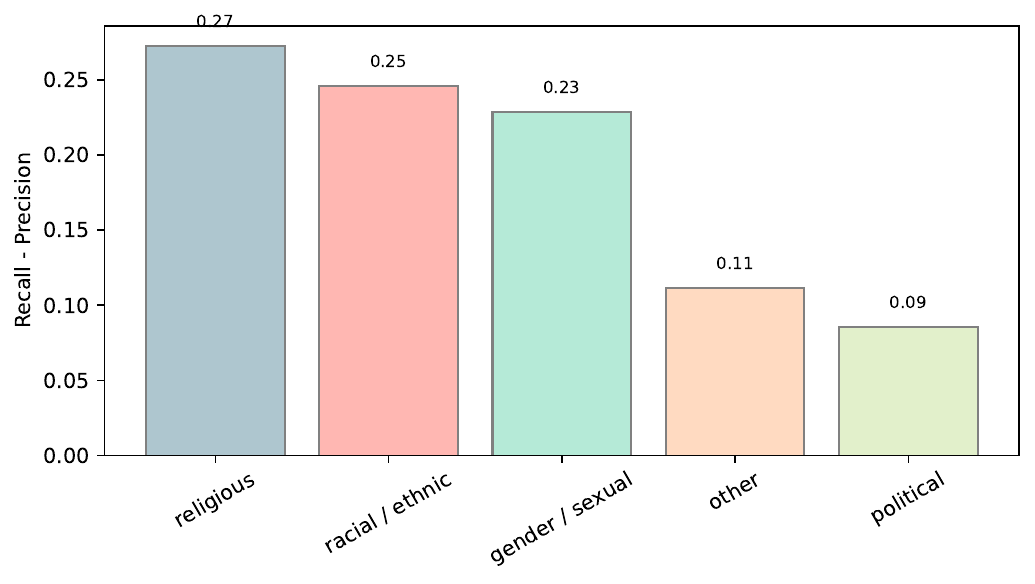}
        \caption{Average precision--recall gap for Subtask~2. Positive values indicate labels for which recall tends to exceed precision, suggesting over-prediction under sparsity.}
        \label{fig:app-s2-pr-gap}
    \end{subfigure}
    \caption{Supplementary post-hoc visualizations for Subtask~2.}
    \label{fig:app-s2}
\end{figure}

Figures~\ref{fig:app-s2-avg-f1} and~\ref{fig:app-s2-pr-gap} provide a label-level view of Subtask~2. Figure~\ref{fig:app-s2-avg-f1} shows that some target types are substantially easier to predict than others, consistent with the coverage--learnability pattern discussed in the main text. Figure~\ref{fig:app-s2-pr-gap} further shows that several labels exhibit recall-dominant behavior, which is consistent with over-prediction under sparse supervision.

\subsection{Additional Analysis for Subtask~3}

\begin{figure}[htbp]
    \centering
    \begin{subfigure}[t]{0.48\textwidth}
        \centering
        \includegraphics[width=\linewidth]{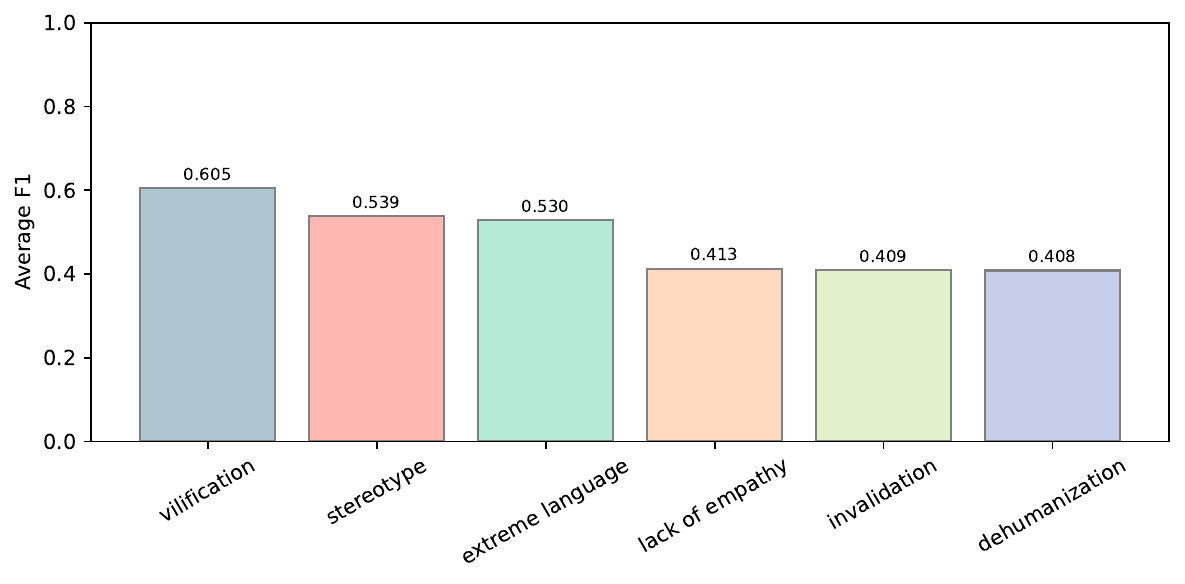}
        \caption{Average label-wise F1 for Subtask~3, illustrating the relative difficulty of different manifestation categories.}
        \label{fig:app-s3-avg-f1}
    \end{subfigure}
    \hfill
    \begin{subfigure}[t]{0.48\textwidth}
        \centering
        \includegraphics[width=\linewidth]{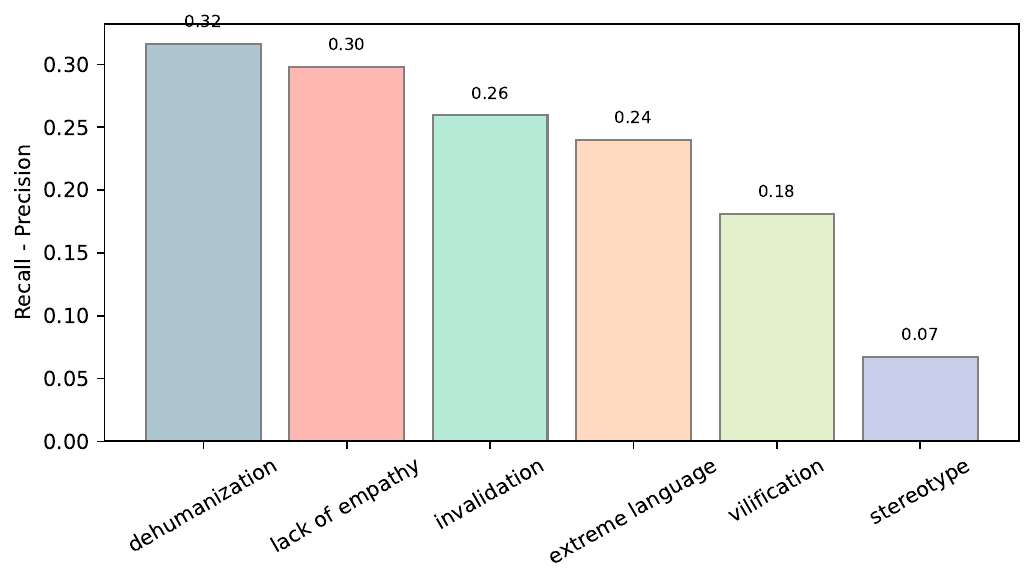}
        \caption{Average precision--recall gap for Subtask~3. Positive values indicate recall-dominant behavior, consistent with miscalibration on sparse labels.}
        \label{fig:app-s3-pr-gap}
    \end{subfigure}
    \caption{Supplementary post-hoc visualizations for Subtask~3.}
    \label{fig:app-s3}
\end{figure}

Figures~\ref{fig:app-s3-avg-f1} and~\ref{fig:app-s3-pr-gap} provide a corresponding label-level view of Subtask~3. Figure~\ref{fig:app-s3-avg-f1} illustrates the relative difficulty of different manifestation categories, while Figure~\ref{fig:app-s3-pr-gap} shows that recall often exceeds precision for the harder labels, supporting the main-text observation that sparse manifestation labels are especially prone to miscalibration.

\end{document}